\documentclass[11pt,a4paper]{article}
\usepackage[hyperref]{acl2019}
\usepackage{times}
\usepackage{latexsym}

\usepackage{url}

\usepackage{graphicx}
\usepackage{amsmath,bm}
\usepackage{mathrsfs}
\usepackage{enumitem}
\aclfinalcopy 
\def\aclpaperid{1868} 

\title{Sequence-to-Nuggets: Nested Entity Mention Detection via \\ Anchor-Region Networks}

\author{Hongyu Lin${}^{1,3}$, Yaojie Lu${}^{1,3}$, Xianpei Han${}^{1,2,*}$, Le Sun${}^{1,2}$ \\
${}^{1}$Chinese Information Processing Laboratory ~ ${}^{2}$State Key Laboratory of Computer Science \\
Institute of Software, Chinese Academy of Sciences, Beijing, China\\
${}^{3}$University of Chinese Academy of Sciences, Beijing, China \\
 {\tt \{hongyu2016,yaojie2017,xianpei,sunle\}@iscas.ac.cn}
}

\date{}

\begin{document}
\maketitle
\begin{abstract}
  Sequential labeling-based NER approaches restrict each word belonging to at most one entity mention, which will face a serious problem when recognizing nested entity mentions. In this paper, we propose to resolve this problem by modeling and leveraging the head-driven phrase structures of entity mentions, i.e., although a mention can nest other mentions, they will not share the same head word. Specifically, we propose \emph{Anchor-Region Networks (ARNs)}, a sequence-to-nuggets architecture for nested mention detection. ARNs first identify anchor words (i.e., possible head words) of all mentions, and then recognize the mention boundaries for each anchor word by exploiting regular phrase structures. Furthermore, we also design \emph{Bag Loss}, an objective function which can train ARNs in an end-to-end manner without using any anchor word annotation. Experiments show that ARNs achieve the state-of-the-art performance on three standard nested entity mention detection benchmarks.

\end{abstract}

\section{Introduction}
{
\renewcommand{\thefootnote}{\fnsymbol{footnote}}
\footnotetext[1]{Corresponding author.}
}
Named entity recognition (NER), or more generally entity mention detection\footnote{In entity mention detection, a mention can be either a named, nominal or pronominal reference of an entity~\cite{N18-1079}.}, aims to identify text spans pertaining to specific entity types such as \emph{Person}, \emph{Organization} and \emph{Location}. NER is a fundamental task of information extraction which enables many downstream NLP applications, such as relation extraction~\cite{guodong2005exploring,mintz2009distant}, event extraction~\cite{ji2008refining,li2013joint} and machine reading comprehension~\cite{rajpurkar2016squad,wang2016multi}.

\begin{figure}
  \setlength{\belowcaptionskip}{-0.4cm}
  \centering
  \includegraphics[width=0.45\textwidth]{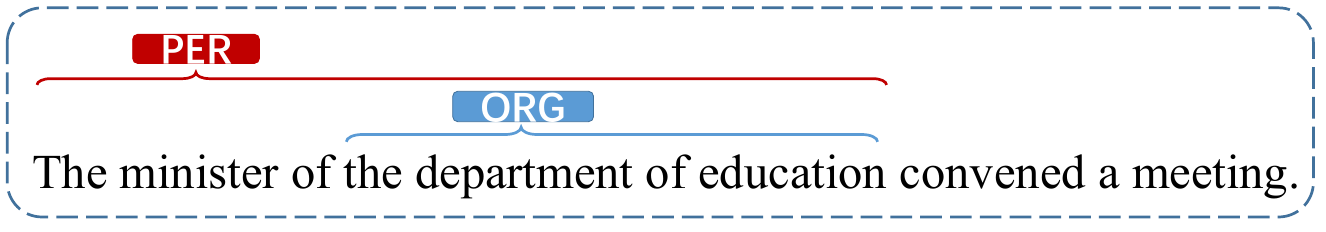}\\
  \caption{An example of nested entity mentions. Due to the nested structure, ``the'',``department'',``of'' and ``education'' belong to both \emph{PER} and \emph{ORG} mentions.}
  \label{fig:exp_nested_mention}
\end{figure}

Previous approaches~\cite{zhou2002named,chieu2002named,bender2003maximum,settles2004biomedical,lample2016neural} commonly regard NER as a sequential labeling task, which generate label sequence for each sentence by assigning one label to each token. These approaches commonly restrict each token belonging to at most one entity mention and, unfortunately, will face a serious problem when recognizing nested entity mentions, where one token may belong to multiple mentions.
For example in Figure~\ref{fig:exp_nested_mention}, an \emph{Organization} entity mention ``the department of education'' is nested in another \emph{Person} entity mention ``the minister of the department of education''. Nested entity mentions are very common.  For instance, in the well-known ACE2005 and RichERE datasets, more than 20\% of entity mentions are nested in other mentions. Therefore, it is critical to consider nested mentions for real-world applications and downstream tasks.

In this paper, we propose a sequence-to-nuggets approach, named as \emph{Anchor-Region Networks (ARNs)}, which can effectively detect all entity mentions by modeling and exploiting the head-driven phrase structures~\cite{pollard1994head,collins2003head} of them. ARNs originate from two observations. First, although an entity mention can nest other mentions, they will not share the same head word. And the head word of a mention can provide strong semantic evidence for its entity type~\cite{P18-1009}. For example in Figure~\ref{fig:exp_nested_mention}, although the \emph{ORG} mention is nested in the \emph{PER} mention, they have different head words ``department'' and ``minister'' respectively, and these head words strongly indicate their corresponding entity types to be \emph{ORG} and \emph{PER}. Second, entity mentions mostly have regular phrase structures. For the two mentions in Figure~\ref{fig:exp_nested_mention}, they share the same ``DET NN of NP'' structure, where the NN after DET are their head words. Based on above observations, entity mentions can be naturally detected in a sequence-to-nuggets manner by 1) identifying the head words of all mentions in a sentence; and 2) recognizing entire mention nuggets centered at detected head words by exploiting regular phrase structures of entity mentions.

\begin{figure}[t]
  \setlength{\belowcaptionskip}{-0.5cm}
  \centering
  \includegraphics[width=0.47\textwidth]{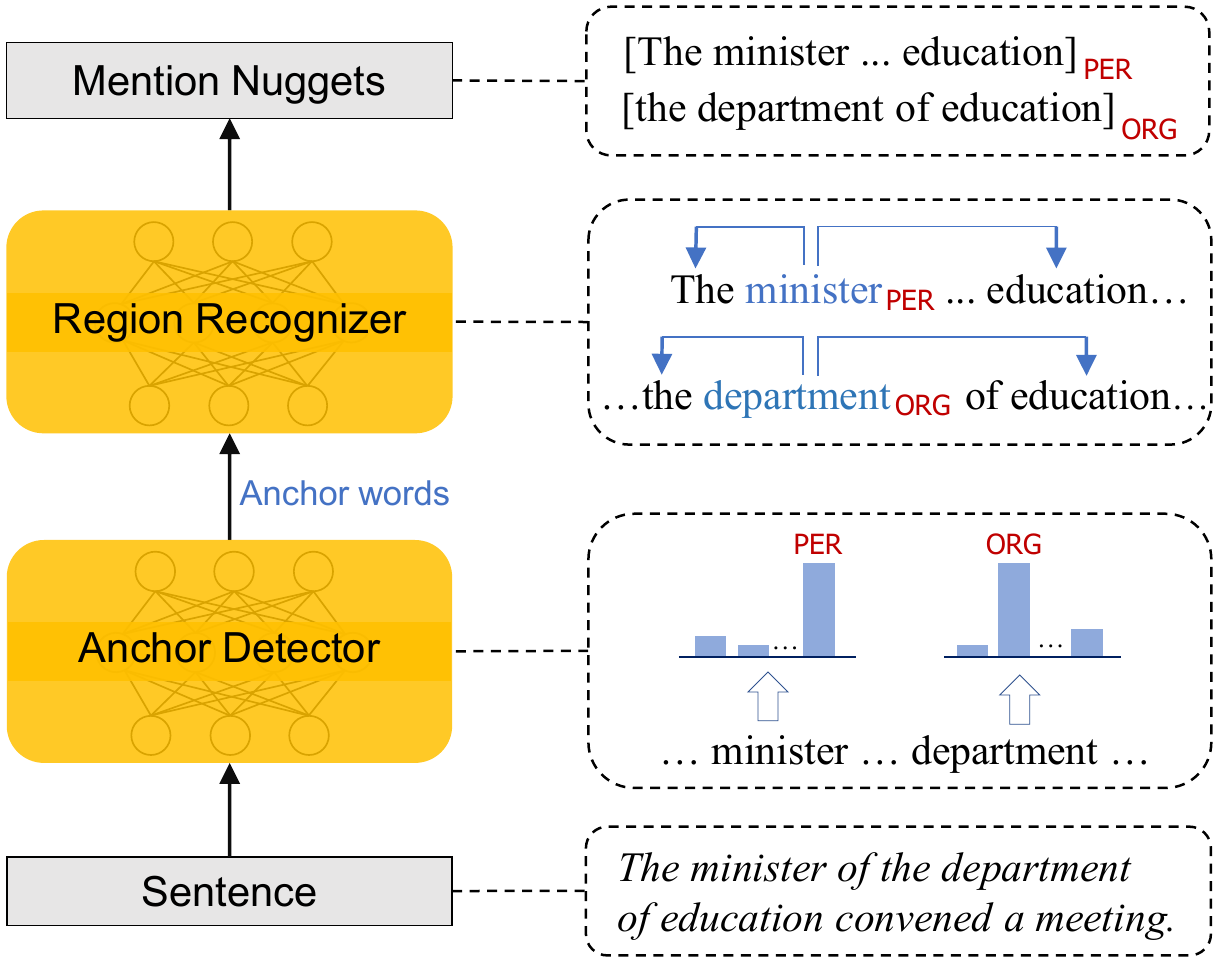}\\
  \caption{The overall architecture of ARNs. Here ``minister'' and ``department'' are detected anchor words for two mentions respectively.}\label{fig:model_archi}
\end{figure}

To this end, we propose ARNs, a new neural network-based approach for nested mention detection. Figure~\ref{fig:model_archi} shows the architecture of ARNs. First, ARNs employs an \emph{anchor detector network} to identify whether each word is a head word of an entity mention, and we refer the detected words as \emph{anchor words}. After that, a \emph{region recognizer network} is used to determine the mention boundaries centering at each anchor word. By effectively capturing head-driven phrase structures of entity mentions, the proposed ARNs can naturally address the nested mention problem because different mentions have different anchor words, and different anchor words correspond to different mention nuggets.

Furthermore, because the majority of NER datasets are not annotated with head words, they cannot be directly used to train our anchor detector. To address this issue, we propose \emph{Bag Loss}, an objective function which can be used to train ARNs in an end-to-end manner without any anchor word annotation. Specifically, our Bag Loss is based on \emph{at-least-one assumption}, i.e., each mention should have at least one anchor word, and the anchor word should strongly indicate its entity type. Based on this assumption, Bag Loss can automatically select the best anchor word within each mention during training, according to the association between words and the entity type of the mention.
For example, given an \emph{ORG} training instance ``the department of education'', Bag Loss will select ``department'' as the anchor word of this mention based on its tight correlation with type \emph{ORG}. While other words in the mention, such as ``the'' and ``of'', will not be regarded as anchor words, because of their weak association with \emph{ORG} type.

We conducted experiments on three standard nested entity mention detection benchmarks, including ACE2005, GENIA and TAC-KBP2017 datasets. Experiments show that ARNs can effectively detect nested entity mentions and achieve the state-of-the-art performance on all above three datasets. For better reproduction, we openly release the entire project at \url{github.com/sanmusunrise/ARNs}.

Generally, our main contributions are:
\begin{itemize}[leftmargin=0.6cm,topsep=0.1cm]
\setlength{\itemsep}{0cm}
\setlength{\parskip}{0.1cm}
    \item We propose a new neural network architecture named as \emph{Anchor-Region Networks}. By effectively modeling and leveraging the head-driven phrase structures of entity mentions, ARNs can naturally handle the nested mention detection problem and achieve the state-of-the-art performance on three benchmarks. To the best of our knowledge, this is the first work which attempts to exploit the head-driven phrase structures for nested NER.
    \item We design an objective function, named as \emph{Bag Loss}. By exploiting the association between words and entity types, Bag Loss can effectively learn ARNs in an end-to-end manner, without using any anchor word annotation.
    \item Head-driven phrase structures are widely spread in natural language. This paper proposes an effective neural network-based solution for exploiting this structure, which can potentially benefit many NLP tasks, such as semantic role labeling~\cite{zhou2015end,he2017deep} and event extraction~\cite{chen2015event,lin2018nugget}.
\end{itemize}

\section{Related Work}
Nested mention detection requires to identify all entity mentions in texts, rather than only outmost mentions in conventional NER. This raises a critical issue to traditional sequential labeling models because they can only assign one label to each token. To address this issue, mainly two kinds of methods have been proposed.

\textbf{Region-based approaches} detect mentions by identifying over subsequences of a sentence respectively, and nested mentions can be detected because they correspond to different subsequences. For this, \citet{finkel2009nested} regarded nodes of parsing trees as candidate subsequences. Recently, \citet{P17-1114} and~\citet{D18-1309} tried to directly classify over all subsequences of a sentence. Besides, \citet{D18-1124} proposed a transition-based method to construct nested mentions via a sequence of specially designed actions. Generally, these approaches are straightforward for nested mention detection, but mostly with high computational cost as they need to classify over almost all sentence subsequences.

\textbf{Schema-based approaches} address nested mentions by designing more expressive tagging schemas, rather than changing tagging units. One representative direction is hypergraph-based methods~\cite{lu2015joint,N18-1079,D18-1019}, where hypergraph-based tags are used to ensure nested mentions can be recovered from word-level tags. Besides, \citet{D17-1276} developed a gap-based tagging schema to capture nested structures. However, these schemas should be designed very carefully to prevent spurious structures and structural ambiguity~\cite{D18-1019}. But  more expressive, unambiguous schemas will inevitably lead to higher time complexity during both training and decoding.

Different from previous methods, this paper proposes a new architecture to address nested mention detection. Compared with region-based approaches, our ARNs detect mentions by exploiting head-driven phrase structures, rather than exhaustive classifying over subsequences. Therefore ARNs can significantly reduce the size of candidate mentions and lead to much lower time complexity. Compared with schema-based approaches, ARNs can naturally address nested mentions since different mentions will have different anchor words. There is no need to design complex tagging schemas, no spurious structures and no structural ambiguity.

Furthermore, we also propose Bag Loss, which can train ARNs in an end-to-end manner without any anchor word annotation. The design of Bag Loss is partially inspired by multi-instance learning (MIL)~\cite{zhou2007multi,zhou2009multi,surdeanu2012multi}, but with a different target. MIL aims to predict a unified label of \emph{a bag of instances}, while Bag Loss is proposed to train ARNs whose anchor detector is required to predict the label of \emph{each instance}. Therefore previous MIL methods are not suitable for training ARNs.

\section{Anchor-Region Networks for Nested Entity Mention Detection}
Given a sentence, Anchor-Region Networks detect all entity mentions in a two-step paradigm. First, an \emph{anchor detector network} identifies anchor words and classifies them into their corresponding entity types. After that, a \emph{region recognizer network} is applied to recognize the entire mention nugget centering at each anchor word. In this way, ARNs can effectively model and exploit head-driven phrase structures of entity mentions: the anchor detector for recognizing possible head words and the region recognizer for capturing phrase structures. These two modules are jointly trained using the proposed Bag Loss, which learns ARNs in an end-to-end manner without using any anchor word annotation. This section will describe the architecture of ARNs. And Bag Loss will be introduced in the next section.

\subsection{Anchor Detector}
An anchor detector is a word-wise classifier, which identifies whether a word is an anchor word of an entity mention of specific types. For the example in Figure~\ref{fig:exp_nested_mention}, the anchor detector should identify that ``minister'' is an anchor word of a \emph{PER} mention and ``department'' is an anchor word of an \emph{ORG} mention.

Formally, given a sentence $x_1,x_2,...,x_n$, all words are first mapped to a sequence of word representations $\bm{x_1},\bm{x_2},...,\bm{x_n}$ where $\bm{x_i}$ is a combination of word embedding, part-of-speech embedding and character-based representation of word $x_i$ following~\citet{lample2016neural}. Then we obtain a context-aware representation $\bm{h^{A}_i}$ of each word $x_i$ using a bidirectional LSTM layer:

\begin{small}
\begin{equation}
  \centering
  \begin{gathered}
    \overrightarrow{\bm{h^{A}_i}}  = {\rm LSTM} (\bm{x_i},\bm{\overrightarrow{h^{A}_{i-1}}}) \\
    \overleftarrow{\bm{h^{A}_i}}  = {\rm LSTM} (\bm{x_i},\bm{\overleftarrow{h^{A}_{i+1}}})\\
    \bm{h^{A}_i} = [\overrightarrow{\bm{h^{A}_i}}  ; \overleftarrow{\bm{h^{A}_i}}] \\
  \end{gathered}
\end{equation}
\end{small}%
The learned representation $\bm{h^{A}_i}$ is then fed into a multi-layer perceptron(MLP) classifier, which computes the scores $\bm{O^{A}_{i}}$ of the word $x_i$ being an anchor word of specific entity types (or \emph{NIL} if this word is not an anchor word):

\begin{small}
\begin{equation}
\bm{O^{A}_{i}}  = {\rm MLP} (\bm{h^{A}_i})
\end{equation}
\end{small}%
where $\bm{O^{A}_{i}} \in R^{|C|}$ and $|C|$ is the number of entity types plus one \emph{NIL} class. Finally a softmax layer is used to normalize $\bm{O^{A}_{i}}$ to probabilities:

\begin{small}
\begin{equation}
P(c_{j}|x_i) = \frac{e^{O^{A}_{ij}}}{\sum_{k=1}^{|C|} e^{O^{A}_{ik}}}
\end{equation}
\end{small}%
where $O^{A}_{ij}$ is the $j^{th}$ element in $\bm{O^{A}_{i}}$, $P(c_{j}|x_i)$ is the probability of word $x_i$ being an anchor word of class $c_j$.
Note that because different mentions will not share the same anchor word, the anchor detector can naturally solve nested mention detection problem by recognizing different anchor words for different mentions.

\subsection{Region Recognizer}
Given an anchor word, ARNs will determine its exact mention nugget using a region recognizer network. For the example in Figure~\ref{fig:exp_nested_mention}, the region recognizer will recognize that ``the minister of the department of education'' is the mention nugget for anchor word ``minister'' and ``the department of education'' is the mention nugget for anchor word ``department''. Inspired by the recent success of pointer networks~\cite{vinyals2015pointer,wang2016machine}, this paper designs a pointer-based architecture to recognize the mention boundaries centering at an anchor word. That is, our region recognizer will detect the mention nugget ``the department of education'' for anchor word ``department'' by recognizing ``the'' to be the left boundary and ``education'' to be the right boundary.

Similar to the anchor detector, a bidirectional LSTM layer is first applied to obtain the context-aware representation $\bm{h^{R}_i}$ of word $x_i$. For recognizing mention boundaries, local features commonly play essential roles. For instance, a noun before a verb is an informative boundary indicator for entity mentions. To capture such local features, we further introduce a convolutional layer upon $\bm{h^{R}_i}$:

\begin{small}
\begin{equation}
\bm{r_{i}} = \tanh(\bm{W} \bm{h^R_{i-k:i+k}} + \bm{b})
\end{equation}
\end{small}%
where $\bm{h^R_{i-k:i+k}}$ is the concatenation of vectors from $\bm{h^R_{i-k}}$ to $\bm{h^R_{i+k}}$, $\bm{W}$ and $\bm{b}$ are the convolutional kernel and the bias term respectively. $k$ is the (one-side) window size of convolutional layer. Finally, for each anchor word $x_i$, we compute its left mention boundary score $L_{ij}$ and right mention boundary score $R_{ij}$ at word $x_j$ by

\begin{small}
\begin{equation}
  \centering
  \begin{gathered}
    L_{ij} = \tanh (\bm{r_j^T \Lambda_1 h^R_i} + \bm{U_1r_j} + b_1) \\
    R_{ij} = \tanh (\bm{r_j^T \Lambda_2 h^R_i} + \bm{U_2r_j} + b_2)\\
  \end{gathered}
\end{equation}
\end{small}%
In the above two equations, the first term within the $\tanh$ function computes the score of word $x_j$ serving as the left/right boundary of a mention centering at word $x_i$. And the second term models the possibility of word $x_j$ itself serving as the boundary universally.
After that, we select the best left boundary word $x_j$ and best right boundary word $x_k$ for anchor word $x_i$, and the nugget \{$x_j,...,x_i,...,x_k$\} will be a recognized mention.

\section{Model Learning with Bag Loss}

\begin{figure*}
  \centering
  \setlength{\belowcaptionskip}{-0.4cm}
  \includegraphics[width=0.8\textwidth]{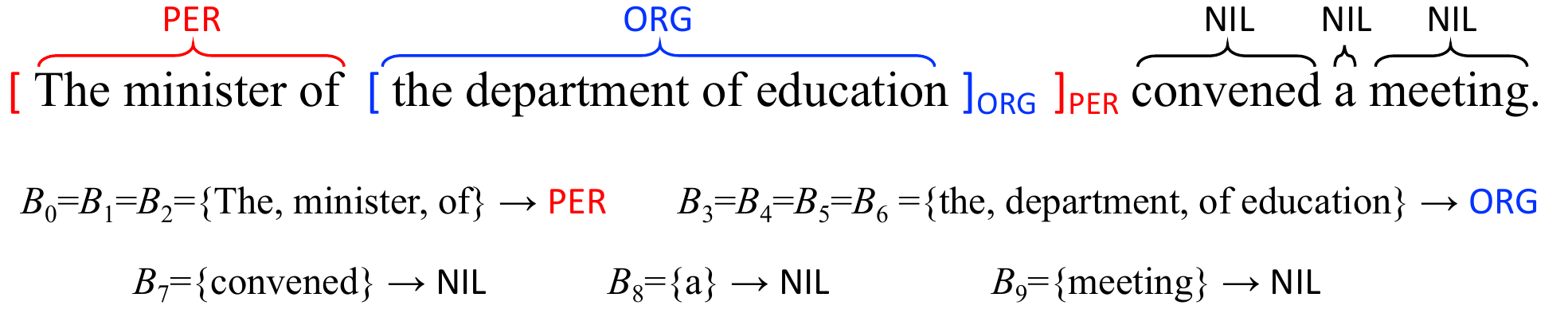}\\
  \caption{An illustration of bags. $B_i$ represents the bag where word $x_i$ is in. This sentence forms five bags, two of which correspond to two entity mentions and three of which correspond to \emph{NIL}.}
  \label{fig:bags}
\end{figure*}

This section describes how to train ARNs using existing NER datasets. The main challenge here is that current NER corpus are not annotated with anchor words of entity mentions, and therefore they cannot be directly used to train the anchor detector. To address this problem, we propose \emph{Bag Loss}, an objective function which can effectively learn ARNs in an end-to-end manner, without using any anchor word annotation.

Intuitively, one naive solution is to regard all words in a mention as its anchor words. However, this naive solution will inevitably result in two severe problems. First, a word may belong to different mentions when nested mentions exist. Therefore this naive solution will lead to ambiguous and noisy anchor words. For the example in Figure~\ref{fig:exp_nested_mention}, it is unreasonable to annotate the word ``department'' as an anchor word of both \emph{PER} and \emph{ORG} mentions, because it has little association to \emph{PER} type although the \emph{PER} mention also contains it. Second, many words in a mention are just function words, which are not associated with its entity type. For example, words ``the'',``of'' and ``education'' in ``the department of education'' are not associated with its type \emph{ORG}. Therefore annotating them as anchor words of the \emph{ORG} mention will introduce remarkable noise.

To resolve the first problem, we observe that a word can only be the anchor word of the innermost mention containing it. This is because a mention nested in another mention can be regarded as a replaceable component, and changing it will not affect the structure of outer mentions. For the case in Figure~\ref{fig:exp_nested_mention}, if we replace the nested mention ``the department of education'' by other \emph{ORG} mention(e.g., changing it to ``State''), the type of the outer mention will not change. Therefore, words in a nested mention should not be regarded as the anchor word of outer mentions, and therefore a word can only be assigned as the anchor word of the innermost mention containing it.

To address the second problem, we design \emph{Bag Loss} based on the \emph{at-least-one assumption}, i.e., for each mention at least one word should be regarded as its anchor word. Specifically, we refer to all words belonging to the same innermost mention as a \emph{bag}. And the type of the bag is the type of that innermost mention.
For example, in Figure~\ref{fig:bags},\{the, minister, of\} will form a \emph{PER} bag, and \{the, department, of education\} will form an \emph{ORG} bag. Besides, each word not covered by any mention will form a one-word bag with \emph{NIL} type. So there are three \emph{NIL} bags in Figure~\ref{fig:bags}, including \{convened\}, \{a\} and \{meeting\}.

Given a bag, Bag Loss will make sure that at least one word in each bag will be selected as its anchor word, and be assigned to the bag type. While other words in that bag will be classified into either the bag type or \emph{NIL}. Bag Loss selects anchor words according to their associations with the bag type. That is, only words highly related to the bag type (e.g., ``department'' in ``the department of education'') will be trained towards the bag type, and other irrelevant words (e.g., ``the'' and ``of'' in the above example) will be trained towards \emph{NIL}.

\noindent \textbf{Bag Loss based End-to-End Learning.} For ARNs, each training instance is a tuple $x = (x_i,x_j,x_k,c_i)$, where $x_j,...,x_k$ is an entity mention with left boundary $x_j$ and right boundary $x_k$. $c_j$ is its entity type and word $x_i$ is a word in this mention's bag\footnote{For words not in any mention, we define $x_j= x_k=x_i$ and $c_i$ = \emph{NIL}, but their boundary will not be considered during optimization according to Equation (\ref{eq:marginloss}).}. For each instance, Bag loss considers two situations: 1) If $x_i$ is its anchor word, the loss will be the sum of the anchor detector loss (i.e., the loss of correctly classifying $x_i$ into its bag type $c_i$) and the region recognizer loss (i.e., the loss of correctly recognizing the mention boundary $x_j$ and $x_k$); 2) If $x_i$ is not its anchor word, the loss will be only the anchor detector loss (i.e., correctly classifying $x_i$ into \emph{NIL}). The final loss for this instance is a weighted sum of the loss of these two situations, where the weight are determined using the association between word $x_i$ and the bag type $c_i$ compared with other words in the same bag. Formally, Bag Loss is written as:

\begin{small}
\begin{equation}
\begin{aligned}
\mathcal{L}(x_i;\theta) =~&\omega_i \cdot [-\log P(c_{i}|x_i) + L^R(x_i;\theta)]\\
&+ (1- \omega_i ) \cdot [-\log P(NIL|x_i)]
\end{aligned}
\label{eq:bagloss}
\end{equation}
\end{small}%
where $- \log P(c_{i}|x_i)$ is the anchor detector loss. $\mathcal{L}^R(x_i;\theta) = \mathcal{L}^{left}(x_i;\theta) +  \mathcal{L}^{right}(x_i;\theta)$ is the loss for the region recognizer measuring how preciously the region recognizer can identify the boundaries centered at anchor word $x_i$. We define $\mathcal{L}^{left}(x_i;\theta)$ using max-margin loss:

\begin{small}
\begin{equation}
\mathcal{L}^{left}(x_i;\theta)=\left\{
\begin{aligned}
&0,&c_i = NIL \\
\max(0,\gamma - & L_{ij} + \max_{t \neq j}L_{it}),&c_i \neq NIL
\end{aligned}
\right.
\label{eq:marginloss}
\end{equation}
\end{small}%
where $\gamma$ is a hyper-parameter representing the margin, and $\mathcal{L}^{right}(x_i;\theta)$ is similarly defined.

Besides, $\omega_i$ in Equation (\ref{eq:bagloss}) measures the correlation between word $x_i$ and the bag type $c_i$. Compared with other words in the same bag, a word $x_i$ should have larger $w_i$ if it has a tighter association with the bag type. Therefore, $\omega_i$ can be naturally defined as:

\begin{small}
\begin{equation}
\omega_i = [\frac{P(c_{i}|x_i)}{\max_{x_t \in B_i} P(c_{i}|x_t)}]^{\alpha}.
\label{eq:weight}
\end{equation}
\end{small}%
where $B_i$ denotes the bag $x_i$ belonging to, i.e., all words that share the same innermost mention with $x_i$. $\alpha$ is a hyper-parameter controlling how likely a word will be regarded as an anchor word rather than regarded as \emph{NIL}. $\alpha=0$ means that all words are annotated with the bag type. And $\alpha \to +\infty$ means that Bag Loss will only choose the word with highest $P(c_i|x_i)$ as anchor word, while all other words in the same bag will be regarded as \emph{NIL}. Consequently, Bag Loss guarantees that at least one anchor word (the one with highest $P(c_{i}|x_i)$, and its corresponding $w_i$ will be 1.0) will be selected for each bag. For other words that are not associated with the type (the ones with low $P(c_{i}|x_i)$), Bag Loss can make it to automatically learn towards \emph{NIL} during training. 

\section{Experiments}

\subsection{Experimental Settings}
We conducted experiments on three standard English entity mention detection benchmarks with nested mentions: ACE2005, GENIA and TAC-KBP2017 (KBP2017) datasets. For ACE2005 and GENIA, we used the same setup as previous work~\cite{N18-1131,D18-1124,D18-1019,N18-1079}. For KBP2017, we evaluated our model on the 2017 English evaluation dataset (LDC2017E55), using previous RichERE annotated datasets (LDC2015E29, LDC2015E68, LDC2016E31 and LDC2017E02) as the training set except 20 randomly sampled documents reserved as development set. Finally, there were 866/20/167 documents for KBP2017 train/dev/test set. In ACE2005, GENIA and KBP2017, there are 22\%, 10\% and 19\% mentions nested in other mentions respectively. We used Stanford CoreNLP toolkit~\cite{manning2014stanford} to preprocess all documents for sentence splitting and POS tagging. Adadelta update rule~\cite{zeiler2012adadelta} is applied for optimization. Word embeddings are initialized with pretrained 200-dimension Glove~\cite{pennington2014glove} vectors\footnote{\url{http://nlp.stanford.edu/data/glove.6B.zip}}. Hyper-parameters are tuned on the development sets\footnote{The hyper-parameter configures are openly released together with our source code at \url{github.com/sanmusunrise/ARNs}.} apart from $\alpha$ in Equation (\ref{eq:weight}), which will be further discussed in Section \ref{sec:ana_bagloss}.

\begin{table*}[!ht]
\begin{center}
\resizebox{0.95\textwidth}{!}{
\begin{tabular}{l||c|c|c||c|c|c||c|c|c||c}
  \hline
   & \multicolumn{3}{c||}{\bf ACE2005} & \multicolumn{3}{c||}{\bf GENIA}  & \multicolumn{3}{c||}{\bf KBP2017}    & \textbf{Time}  \\ \cline{2-10}
  Model                                   & P             & R         & F1    & P            & R         & F1    & P              & R          & F1 & \textbf{Complexity}    \\ \hline

  LSTM-CRF~\cite{lample2016neural}        & 70.3          & 55.7      & 62.2  & 75.2         & 64.6      & 69.5  & 71.5           & 53.3       & 61.1  & $O(mn)$   \\
  Multi-CRF                               & 69.7          & 61.3      & 65.2  & 73.1         & 64.9      & 68.8  & 69.7           & 60.8       & 64.9  & $O(mn)$    \\ \hline

  FOFE(c=6)~\cite{P17-1114}               & 76.5          & 66.3      & 71.0  & 75.4         & 67.8      & 71.4  & 81.8           & 62.0       & 70.6     & $O(mn^2)$   \\
  FOFE(c=n)~\cite{P17-1114}               & 76.9          & 62.0      & 68.7  & 74.0         & 65.5      & 69.5  & 79.1           & 62.5       & 69.8  & $O(mn^2)$    \\
  Transition~\cite{D18-1124}              & 74.5          & 71.5      & 73.0  & 78.0         & 70.2      & 73.9  & 74.7           & 67.0       & 70.1  & $O(mn)$    \\
  Cascaded-CRF~\cite{N18-1131}            & 74.2          & 70.3      & 72.2  & 78.5         & 71.3      & 74.7  & -              & -          & -     & -    \\ \hline

  LH~\cite{N18-1079}                      & 70.6          & 70.4      & 70.5  & 79.8         & 68.2      & 73.6  & -              & -          & -     & $O(mn)$  \\
  SH(c=6)~\cite{D18-1019}                 & 75.9          & 70.0      & 72.8  & 76.8         & 71.8      & 74.2  & 73.3           & 65.8       & 69.4  & $O(cmn)$ \\
  SH(c=n)~\cite{D18-1019}                 & 76.8          & 72.3      & 74.5  & 77.0         & 73.3      & \textbf{75.1}  & 79.2           & 66.5      & 72.3 & $O(mn^2)$  \\  \hline

  KBP2017 Best~\cite{ji2017overview}      & -             & -         & -     & -            & -         & -     & 72.6           & 73.0       & 72.8   & -  \\  \hline\hline

  Anchor-Region Networks (c=6)            & 75.2          & 72.5      & 73.9  & 75.2         & 73.3      & 74.2  & 76.2           & 71.5       & 73.8   & $O(mn+ck)$   \\
  Anchor-Region Networks (c=n)            & 76.2          & 73.6      & \textbf{74.9}  & 75.8         & 73.9      & 74.8  & 77.7           & 71.8       & \textbf{74.6} & $O(mn+nk)$   \\ \hline
\end{tabular}
}
\end{center}
\caption{Overall experiment results on ACE2005, GENIA and KBP2017 datasets. $c$ is the maximum length of mention and $n$ refers to the length of sentence. For time complexity, $m$ denotes the number of class and $k$ denotes the average number of anchor words in each sentence($k<<n$). The time complexity of Cascaded-CRF depends on datasets so is not listed here.}
\label{tab:overall_rst}
\end{table*}

\subsection{Baselines}
We compare ARNs with following baselines\footnote{As~\citet{D18-1019} reported, neural network-based baselines significantly outperform all non-neural methods. So we only compared with neural network-based baselines.}:

\begin{itemize}[leftmargin=0.3cm,topsep=0.0cm]
\setlength{\itemsep}{0pt}
\setlength{\parskip}{2pt}
\item \textbf{Conventional CRF models}, including \emph{LSTM-CRF}~\cite{lample2016neural} and \emph{Multi-CRF}. LSTM-CRF is a classical baseline for NER, which doesn't consider nested mentions so only outmost mentions are used for training. Multi-CRF is similar to LSTM-CRF but learns one model for each entity type, and thus is able to recognize nested mentions if they have different types.

\item \textbf{Region-based methods}, including \emph{FOFE}~\cite{P17-1114}, \emph{Cascaded-CRF}~\cite{N18-1131} and a transition model (refered as \emph{Transition}) proposed by~\citet{D18-1124}. FOFE directly classifies over all sub-sequences of a sentence and thus all potential mentions can be considered. Cascaded-CRF uses several stacked CRF layers to recognize nested mentions at different levels. Transition constructs nested mentions through a sequence of actions.

\item \textbf{Hypergraph-based methods}, including the \emph{LSTM-Hypergraph (LH)} model~\cite{N18-1079} and the \emph{Segmental Hypergraph (SH)} by~\citet{D18-1019}. LH used an LSTM model to learn features and then decode them into a hypergraph. SH further considered the transition between labels to alleviate labeling ambiguity, which is the state-of-the-art in both ACE2005 and GENIA\footnote{Even~\citet{D18-1309} reported a higher performance on GENIA, their experimental settings are obviously different from other baselines. As they didn't release their dataset splits and source code, we are unable to compare it with listed baselines.} datasets.

\end{itemize}

Besides, we also compared the performance of ARNs with the best system in TAC-KBP 2017 Evaluation~\cite{ji2017overview}. The same as all previous studies, models are evaluated using micro-averaged Precision(P), Recall(R) and F1-score. To balance time complexity and performance, \citet{D18-1019} proposed to restrict the maximum length of mentions to 6, which covers more than 95\% mentions. So we also compared to baselines where the maximum length of mention is restricted or unrestricted. Besides, we also compared the decoding time complexity of different methods.

\subsection{Overall Results}
Table~\ref{tab:overall_rst} shows the overall results on ACE2005, GENIA and KBP2017 datasets. From this table, we can see that:

1) \textbf{Nested mentions have a significant influence on NER performance and are required to be specially treated.} Compared with LSTM-CRF and Multi-CRF baselines, all other methods dealing with nested mentions achieved significant F1-score improvements. So it is critical to take nested mentions into consideration for real-world applications and downstream tasks.

2) \textbf{Our Anchor-Region Networks can effectively resolve the nested mention detection problem, and achieved the state-of-the-art performance in all three datasets.} On ACE2005 and GENIA, ARNs achieved the state-of-the-art performance on both the restricted and the unrestricted mention length settings. On KBP2017, ARNs outperform the top-1 system in the 2017 Evaluation by a large margin. This verifies the effectiveness of our new architecture.

3) \textbf{By modeling and exploiting head-driven phrase structure of entity mentions, ARNs reduce the computational cost significantly.} ARNs only detect nuggets centering at detected anchor words. Note that for each sentence, the number of potential anchor words $k$ is significantly smaller than the sentence length $n$. Therefore the computational cost of our region recognizer is significantly lower than that of traditional region-based methods which perform classification on all sub-sequences, as well as hypergraph-based methods which introduced structural dependencies between labels to prevent structural ambiguity~\cite{D18-1019}. Furthermore, ARNs are highly parallelizable if we replace the BiLSTM context encoder with other parallelizable context encoder architecture (e.g., Transformer~\cite{NIPS2017_7181}).

\begin{table*}
\begin{center}
\begin{tabular}{|c|c|}
\hline
\textbf{Type} & \textbf{Most Frequent Anchor Words }\\ \hline
PER & I, you, he, they, we, people, president, Mandela, family, officials \\ \hline
ORG & government, Apple, they, its, Nokia, company, Microsoft, military, party, bank \\ \hline
FAC & building, home, prison, house, store, factories, factory, school, streets, there \\ \hline
GPE & country, China, U.S., US, Cyprus, our, state, countries, Syria, Russia \\ \hline
LOC & world, moon, areas, space, European, Europe, area, region, places, border \\ \hline \hline
NIL & the, a, of, 's, in, and, to, his, who, former \\ \hline
\end{tabular}
\end{center}
\caption{The top-10 most frequent anchor words of each type on KBP2017 datasets. Line \emph{NIL} shows most frequent words that appears in a mention but are not regarded as anchor words.}
\label{tab:frequent_head_words}
\end{table*}

\subsection{Effects of Bag Loss}
\label{sec:ana_bagloss}

\begin{figure}
  \centering
  \setlength{\belowcaptionskip}{-0.5cm}
  \includegraphics[width=0.45\textwidth]{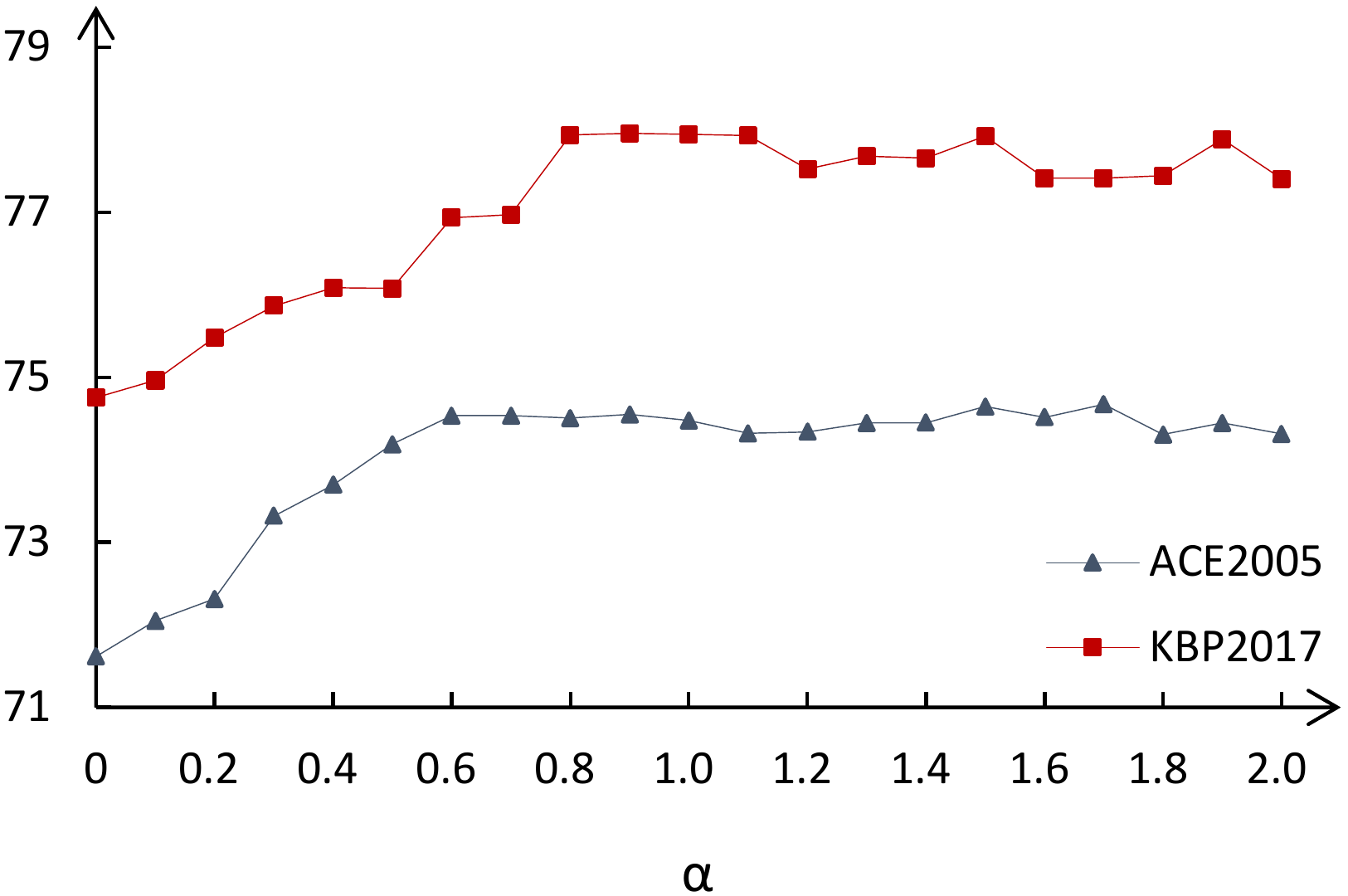}\\
  \caption{The F1-score w.r.t. different $\alpha$ in Bag Loss on development sets. When $\alpha=0$, the model ablates Bag Loss and will treat all words in the same innermost mention as anchor words during training.}
  \label{fig:alpha}
\end{figure}
In this section, we investigate effects of Bag Loss by varying the values of hyper-parameter $\alpha$ in Equation (\ref{eq:weight}) on the system performance. Figure~\ref{fig:alpha} shows the F1 curves on both ACE2005 and KBP2017 datasets when $\alpha$ varies. We can see that:

1) \textbf{Bag Loss is effective for anchor word selection during training.} In Figure~\ref{fig:alpha}, setting $\alpha$ to 0 significantly undermines the performance. Note that setting $\alpha$ to 0 is the same as ablating Bag Loss, i.e., the model will treat all words in the same innermost mention as anchor words. This result further verifies the necessity of Bag Loss. That is, because not all words in a mention are related to its type, it will introduce remarkable noise by regarding all words in mentions as anchor words.

2) \textbf{Bag Loss is not sensitive to $\bm \alpha$ when it is larger than a threshold.} In Figure~\ref{fig:alpha}, our systems achieve nearly the same performance when  $\alpha>0.8$. We find that this is because our model can predict anchor word in a very sharp probability distribution, so slight change of $\alpha$ does not make a big difference. Therefore, in all our experiments we empirically set $\alpha=1$ without special declaration. This also verified that Bag Loss can discover head-driven phrase structure steadily without using anchor word annotations.

\subsection{Further Discussion on Bag Loss and Marginalization-based Loss}
One possible alternative solution for Bag Loss is to regard the anchor word as a hidden variable, and obtain the likelihood of each mention by marginalizing over all words in the mention nugget with

\begin{small}
\begin{equation}
P(c,x_j,x_k) = \sum_{x_i} P(x_i,c)P(x_j,x_k|x_i,c).
\end{equation}
\end{small}%
For $P(x_i,c)$, if we assume that the prior for each word being the anchor word is equal, it can be refactorized by

\begin{small}
\begin{equation}
P(x_i,c) = P(c|x_i)P(x_i) \propto P(c|x_i).
\end{equation}
\end{small}%

However, we find that this approach does not work well in practice. This may because that, as we mentioned above, the prior probability of each word being the anchor word should not be equal. Words with highly semantic relatedness to the types are more likely to be the anchor word. Furthermore, this marginalization-based training object can only guarantee that words being regarded as the  anchor words are trained towards the mention type, but will not encourage the other irrelevant words in the mention to be trained towards \emph{NIL}. Therefore, compared with Bag Loss, the marginalization-based solution can not achieve the promising results for ARNs training.

\subsection{Analysis on Anchor Words}

To analyze the detected anchor words, Table~\ref{tab:frequent_head_words} shows the most common anchor words for all entity types. Besides, words that frequently appear in a mention but being recognized as \emph{NIL} are also presented. We can see that the top-10 anchor words of each type are very convincing: all these words are strong indicators of their entity types. Besides, we can see that frequent \emph{NIL} words in entity mentions are commonly function words, which play significant role in the structure of mention nuggets (e.g., ``the'' and ``a'' often indicates the start of an entity mention) but have little semantic association with entity types. This supports our motivation and further verifies the effectiveness of Bag Loss for anchor word selection.

\subsection{Error Analysis}

\begin{table}
\begin{center}
\resizebox{0.48\textwidth}{!}{
\begin{tabular}{c|c|c|c}
\hline
                          & \textbf{ACE2005} & \textbf{GENIA} & \textbf{KBP2017}         \\ \hline
Anchor Detector           &  82.9         & 82.7           &  83.0                   \\
Entire ARNs               &  74.9         & 74.8           &  74.6                    \\ \hline
$\Delta$                  &  8.0          &  7.9           &  8.4                    \\ \hline
\end{tabular}
}
\end{center}
\caption{F1-scores gap between the anchor detector and the entire ARNs (anchor + region). }
\label{tab:error_analysis_part}
\end{table}

\begin{figure}
\setlength{\belowcaptionskip}{-0.4cm}
  \centering
  \includegraphics[width=0.45\textwidth]{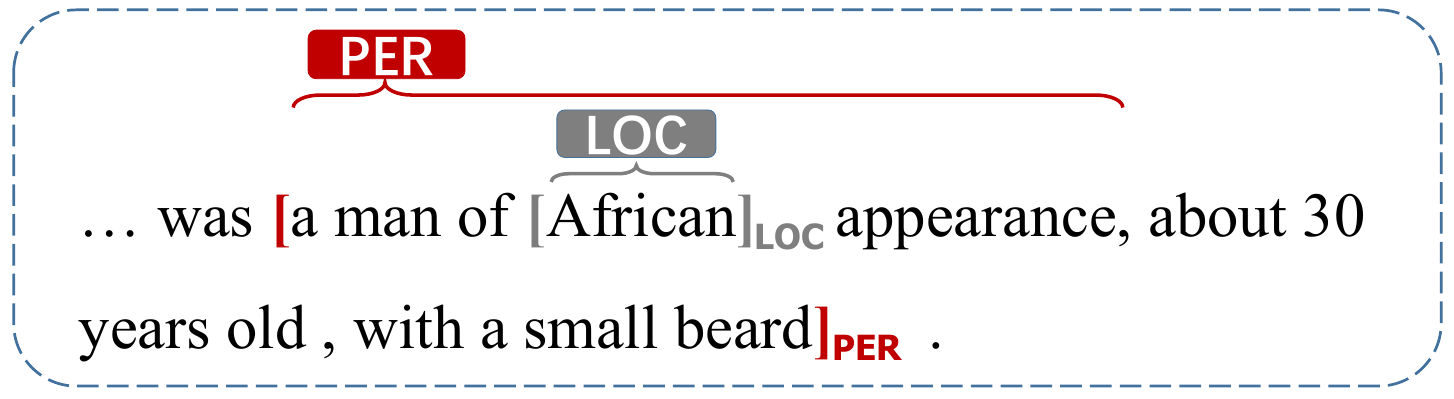}\\
  \caption{A representative error case of ARNs, where the right boundary of the \emph{PER} mention is misclassified. Braces above the sentence indicate the output of ARNs, and brackets in the sentence represent the golden annotation. We find that the majority of errors occur because of the long-term dependencies stemming from postpositive attributive and attributive clauses.}
  \label{fig:err_case}
\end{figure}

This section conducts error analysis on ARNs. Table~\ref{tab:error_analysis_part} shows the performance gap between the anchor detector and the entire ARNs. We can see that there is still a significant performance gap from the anchor detector to entire ARNs. That is, there exist a number of mentions whose anchor words are correctly detected by the anchor detector but their boundaries are mistakenly recognized by the region recognizer. To investigate the reason behind this above performance gap, we analyze these cases and find that most of these errors stem from the existence of postpositive attributive and attributive clauses. Figure~\ref{fig:err_case} shows an error case stemming from postpositive attributive. These cases are quite difficult for neural networks because long-term dependencies between clauses need to be carefully considered. One strategy to handle these cases is to introduce syntactic knowledge, which we leave as future work for improving ARNs.

\section{Conclusions and Future Work}
This paper proposes Anchor-Region networks, a sequence-to-nuggets architecture which can naturally detect nested entity mentions by modeling and exploiting head-driven phrase structures of entity mentions. Specifically, an anchor detector is first used to detect the anchor words of entity mentions and then a region recognizer is designed to recognize the mention boundaries centering at each anchor word. Furthermore, we also propose Bag Loss to train ARNs in an end-to-end manner without using any anchor word annotation. Experiments show that ARNs achieve the state-of-the-art performance on all three benchmarks.

As the head-driven structures are widely spread in natural language, the solution proposed in this paper can also be used for modeling and exploiting this structure in many other NLP tasks, such as semantic role labeling and event extraction.

\section*{Acknowledgments}

We sincerely thank the reviewers for their insightful comments and valuable suggestions. Moreover, this work is supported by the National Natural Science Foundation of China under Grants no. 61433015, 61572477 and 61772505; the Projects of the Chinese Language Committee under Grants no. WT135-24; and the Young Elite Scientists Sponsorship Program no. YESS20160177.

\bibliography{ARNs}
\bibliographystyle{acl_natbib}

\end{document}